\documentclass[runningheads]{llncs}
\usepackage[T1]{fontenc}
\usepackage{bm}
\usepackage{graphicx,verbatim}
\usepackage{amsmath}
\usepackage{booktabs}
\usepackage{array}
\usepackage{graphicx}
\usepackage{subcaption} 
\usepackage{amssymb}
\usepackage[pagebackref=true,breaklinks=true,colorlinks,bookmarks=false]{hyperref}
\usepackage{cleveref}

\usepackage{color}

\begin{document}
\title{CXR-CML: Improved zero-shot classification of long-tailed multi-label diseases in Chest X-Rays} 
 \titlerunning{CXR-CML} 
%

\author{
Rajesh Madhipati\inst{1}\and
Sheethal Bhat\inst{1}\and
Lukas Buess\inst{1} \and
Andreas Maier\inst{1}
}
\authorrunning{R. Madhipati et al.}
\institute{
Friedrich-Alexander University Erlangen-Nuremberg, Erlangen 91054, Germany \\
\email{\{Rajesh.madhipati, Sheethal.bhat, lukas.buess, Andreas.maier\}@fau.de}
}
    
\maketitle              
%
\begin{abstract}
Chest radiography (CXR) plays a crucial role in the diagnosis of various diseases. However, the inherent class imbalance in the distribution of clinical findings presents a significant challenge for current self-supervised deep learning models. These models often fail to accurately classify long-tailed classes. Current Vision-Language models such as Contrastive Language Image Pre-training (CLIP) models effectively model the manifold distribution of the latent space, enabling high zero-shot classification accuracies. Although CLIP performs well on most of the primary classes in the dataset, our work reveals that its effectiveness decreases significantly for classes with a long-tailed distribution. Our approach employs a class-weighting mechanism that directly aligns with the distribution of classes within the latent space. This method ensures a substantial improvement in overall classification performance, with particular emphasis on enhancing the recognition and accuracy of rarely observed classes. We accomplish this by applying Gaussian Mixture Model (GMM) clustering to the latent space. The subsequent clusters are further refined by Student t-distribution, followed by a metric loss that utilizes the altered embeddings. Our approach facilitates stable and adaptive clustering of the features, resulting in a notable average improvement of 2\% in AUC scores across 40 classes in the MIMIC-CXR-JPG dataset from previous State of the ART models(SOTA). \\
Our code is publicly available at: \href{https://anonymous.4open.science/r/CXR_CML-F703 }{CXR-CML} . 
\end{abstract}

\section{Introduction}

Chest radiography (CXR) is one of the most widely utilized diagnostic tools in clinical practice, providing essential information on a variety of pulmonary and cardiothoracic conditions \cite{speets2006chestradiography}. The availability of large public CXR datasets \cite{cxrdatasets} with corresponding clinical reports has driven the development of vision-language (VL) models in CXR Artificial Intelligence research \cite{bannur2024maira}. Moreover the high cost of annotations, coupled with a shortage of radiologists, has prompted investigations on the effective use of self-supervised learning (SSL) methods.  \cite{seputis2024multimodaladaptervisionlanguagemodels,wu2023medklipmedicalknowledgeenhanced,wang2022medclipcontrastivelearningunpaired,mu2021slipselfsupervisionmeetslanguageimage,Delitzas_2023_BMVC}. 
While SSL methods rely on contrastive learning principles, CLIP \cite{clip} directly contrasts the extracted visual and language features, thus achieving impressive zero-shot performance on inherently derived categories. Unlike other SSL methods, CLIP eliminates the need for further category based finetuning. 
Due to the prevalent data distribution in CXR datasets \cite{Park2022},  CLIP based VL-SSL models deliver impressive performance \cite{wang2023clipmed,li2023clipcxr,wu2023medklipmedicalknowledgeenhanced,bannur2024maira} on commonly occuring diseases such as pneumonia or pleural effusion. However, many clinically relevant findings are underrepresented in training distributions \cite{johnson2019mimiccxrjpglargepubliclyavailable}, thus negatively impacting the robustness and clinical applicability of these models \cite{Zhang2023}. Early VL approaches like ConVIRT \cite{zhang2020contrastive} and GLoRIA \cite{huang2021gloria} introduced contrastive learning frameworks to align medical images and text, thereby improving multimodal representation learning. Building on these, CXR-BERT \cite{boecking2022cxrbert} specialized in pretraining on chest X-ray reports, while MedKLIP \cite{wu2023medklipmedicalknowledgeenhanced} enhanced performance by integrating structured clinical knowledge. Recent advancements \cite{cxr_clip,Du_CLEFT_MICCAI2024,cxrirgen}, have further improved zero-shot pathology classification across multiple CXR datasets, demonstrating the reliability of VL models. 

Although notable, these systems remain unsuitable for practical deployment due to inconsistent performance across all categories of pathologies as seen in our previous work [anon]. Specifically, these methods often assume uniform Gaussian distributions in the latent space, which may not adequately represent the distribution heterogeneity present in medical datasets. This leads to suboptimal clustering and conflating feature representation for rare diseases \cite{Holste_2022}.


Consequently, we present CXR-CML (Chest X-ray Contrastive Metric Learning) which seeks to model the latent distribution such that the long-tailed classes are appropriately clustered. For this purpose we first apply a Gaussian Mixture Model (GMM)\cite{gmm} on the latent space derived from CLIP \cite{clip}. A Student t-distribution \cite{student} further refines the GMM \cite{gmm}, enhancing the inherent clusters \cite{clip}. Subsequently, we utilize domain-specific metric learning to leverage these clusters and enhance the feature space, guaranteeing that the model acquires distinct representations for both frequently and rarely seen classes. To evaluate the robustness of our model, we evaluated it in the 40 categories released by the MICCAI challenge \cite{challenge}. This encompasses 12 rare and 28 common classes, offering a comprehensive model evaluation previously unavailable. 

\paragraph{\textbf{Main contributions:}}
    1) We introduce CXR-CML, a method to model the latent distribution manifold more effectively. This is accomplished through the application of GMM \cite{gmm} refined with a Student-t distribution. 2) We leverage the clustered distribution to apply a metric loss that yields robust improvement across a wide range of categories.  
    3) We conduct a robust evaluation using 5-fold cross-validation on 20\% of the training set across 12 long-tailed and 28 base classes in MIMIC-CXR-JPG \cite{johnson2019mimiccxrjpglargepubliclyavailable} . To our knowledge, this study covers the widest range of categories for CXR zero-shot classification evaluation.

\section{Method}

\begin{figure}[t]
    \centering
    \includegraphics[width=1\linewidth]{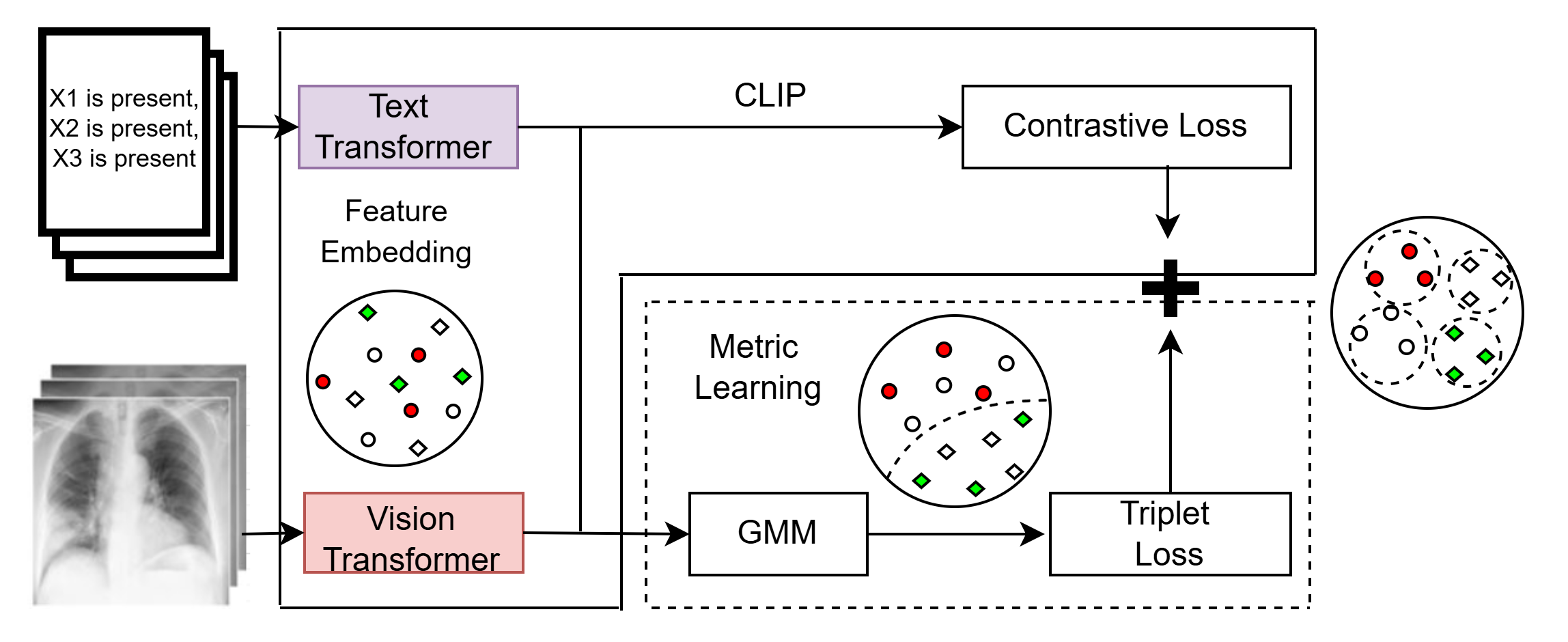}
    \caption{A Text Transformer \& Vision Transformer extract embeddings from textual and visual inputs. A Gaussian Mixture Model (GMM), enhanced by the Student’s t-distribution, is used to form and refine initial clusters, where Label A and Label B represent semantically \& visually similar conditions. Label C \& Label D form another related cluster. We use contrastive and triplet loss, further refining the embedding space by improving intra-class compactness and inter-class separation. For example, Label A \& Label B correspond to 'Atelectasis' and 'Lobar Atelectasis,' while Label C \& Label D correspond to 'Pleural Effusion' and 'Pulmonary Edema,' respectively.}
    \label{fig:framework}
\end{figure}

Given its remarkable zero-shot performance, our method is based on CLIP \cite{radford2021learningtransferablevisualmodels}. The model \cite{radford2021learningtransferablevisualmodels} learns to correlate the images with their corresponding text captions through a cosine similarity loss in a shared latent space. In \textbf{CXR-CML}, we apply GMM \cite{gmm} on the CLIP extracted visual-language embeddings and refine them using a Student t-distribution. This is followed by a metric loss that further refines the feature space. 

\subsubsection{Modeling the latent space: GMM}

The GMM \cite{gmm} is a probabilistic framework that models complex data distributions using multiple Gaussian components. The unsupervised algorithm effectively clusters high-dimensional data, without overfitting to the dominant classes. However, its clusters are soft assignments with overlapping boundaries, making it less distinct than more specialized clustering methods. This makes it especially suitable for multi-label CXR data where images contain multiple pathologies. Thererfore, GMM’s flexibility in approximating diverse distributions makes it a viable choice for modeling heterogeneous CXR data. The algorithm requires us to set the expected number of clusters, denoted as \( N \). We apply GMM only on the visual features, as they exhibit greater variability within our dataset when compared to text. 

\subsubsection{Modeling the distribution: Student t-distribution}

The t-distribution’s \cite{student,ref2} heavy-tailed nature allows it to assign non-negligible probabilities to data points far from the mean, which is critical for capturing underrepresented classes in long-tailed distributions. This property is mathematically supported by its polynomial decay, as opposed to the exponential decay of the Gaussian distribution, enabling it to better model the rare but significant instances often found in medical data. Unlike traditional GMMs \cite{gmm}, which assume Gaussian-distributed data, the t-distribution is better suited for capturing the heavy-tailed nature of medical data, providing more robust covariance estimation and preventing overfitting to Gaussian assumptions. By integrating the t-distribution into our framework, we enhance the model’s ability to handle long-tailed classes, ensuring stable and discriminative clustering. 

Given a batch of feature embeddings \( \mathbf{z} \in \mathbb{R}^d \), we model the data using a mixture of Student’s t-distributions. This is mathematically described as:

\begin{equation}  
p(\mathbf{z}) = \sum_{k=1}^{K} \pi_k \, \mathcal{T}(\mathbf{z} \mid \bm{\mu}_k, \bm{\Sigma}_k, \nu),
\end{equation}  

where \( \pi_k \) is the mixture weight, \( \bm{\mu}_k \) is the mean vector, \( \bm{\Sigma}_k \) is the covariance matrix, and \( \nu \) is the degrees of freedom for the \( k \)-th t-distribution component. The degrees of freedom \( \nu \) control the heaviness of the tails, allowing the model to better adapt to outliers and rare classes.

The Student’s t-distribution is defined as:

\begin{equation}  
\mathcal{T}(\mathbf{z} \mid \bm{\mu}_k, \bm{\Sigma}_k, \nu) = \frac{\Gamma\left(\frac{\nu + d}{2}\right)}{\Gamma\left(\frac{\nu}{2}\right) (\nu \pi)^{d/2} |\bm{\Sigma}_k|^{1/2}} \left(1 + \frac{(\mathbf{z} - \bm{\mu}_k)^T \bm{\Sigma}_k^{-1} (\mathbf{z} - \bm{\mu}_k)}{\nu}\right)^{-\frac{\nu + d}{2}},
\end{equation}  

where \( d \) is the dimensionality of the embeddings, and \( \Gamma(\cdot) \) is the gamma function. As \( \nu \to 1 \), the distribution approaches the standard Cauchy distribution \cite{CauchyDistribution} with heavy tails, while as \( \nu \to \infty \), it converges to the Gaussian distribution. This formulation allows the model to assign higher probabilities to outliers, making it more robust to rare classes. 
As a result, this algorithm enhances existing image-text correlation clusters. 

\subsubsection{Metric Learning: Triplet Loss}

In addition to the original contrastive loss $\mathcal{L}_c$ \cite{clip}, we apply a metric loss on the GMM clusters. Metric learning includes various loss functions, such as Ranked List Loss and center loss, to enhance class discriminability. It is typically used to train a network to distinguish features that are hard to differentiate. In this context, given the data distribution and clusters formed by the GMM phase, we find triplet loss $\mathcal{L}_m$ to be suitable \cite{PCCT,E&E}. The GMM clustering assignments are used to generate pseudo-labels, which guide the selection of triplets needed for $\mathcal{L}_m$. The triplets \( (\mathbf{a}, \mathbf{p}, \mathbf{n}) \) are selected, where \( \mathbf{a} \) is an anchor, \( \mathbf{p} \) is a positive sample (from the same cluster as \( \mathbf{a} \)), and \( \mathbf{n} \) is a negative sample (from a different cluster). The loss is mathematically defined as:

\begin{equation}
\mathcal{L}_m = \sum_{(\mathbf{a}, \mathbf{p}, \mathbf{n})} \max(0, d(\mathbf{a}, \mathbf{p}) - d(\mathbf{a}, \mathbf{n}) + \alpha),
\end{equation}

where \( d(\cdot, \cdot) \) is the standard Euclidean distance between embeddings \( \mathbf{a} \) and \( \mathbf{p} \), and \( \alpha \) is a hyperparameter that defines the margin, ensuring sufficient separation between clusters. While $\mathcal{L}_c$ implicitly encourages intra-cluster compactness and inter-cluster separation, $\mathcal{L}_m$ provides explicit further guidance to the network. The network is trained with the complete loss which is given as,
\begin{equation}
\mathcal{L} = \mathcal{L}_c + \mathcal{L}_m.
\end{equation}

\subsubsection{Text Generation}\label{sec:text-gen}
MIMIC-CXR-JPG \cite{johnson2019mimiccxrjpglargepubliclyavailable} is a multi-label dataset. We generate ``meta labels'' using the NLP generated data provided by MICCAI challenge \cite{challenge}. These labels indicate the absence or presence of specific diseases for each image. Textual descriptions are constructed for all classes that are marked as present in the groundtruth annotations. For example, for an image containing \textit{adenopathy} and \textit{pulmonary effusion}, the generated text is:

\begin{quote}
\textit{``adenopathy is present, pulmonary effusion is present''}
\end{quote}

By utilizing standard disease names, our method enables the text embeddings to act as weak supervisory signals to enhance training.

\section{Experimental Setup}

\subsubsection{Dataset}

We evaluate our method on the MIMIC-CXR-JPG dataset \cite{johnson2019mimiccxrjpglargepubliclyavailable,challenge}, which consists of 234,800 frontal-view chest X-ray JPG images labeled with 39 disease classes. The original MIMIC-CXR-JPG dataset includes 14 disease classes, while the MICCAI challenge \cite{challenge}, titled "CXR-LT: Long-tailed, multi-label, and zero-shot classification on chest X-rays," introduced an additional 26 classes. \cite{challenge} expands the scope of the dataset to include a much larger category of critical and underrepresented classes. Table \ref{tab:class-distribution} denotes the high data imbalance, showing the distribution of disease classes according to the sample count for each category. There are 11 disease classes containing more than 10,000 samples, 17 classes ranging between 1,000 and 10,000 samples, and 12 classes having fewer than 1,000 samples. For our experiments we categorize the dataset into base and rare classes: classes with fewer than 1000 samples are rare and the remaining are base. Consequently, rare classes constitute only 2\% of the dataset. 

\begin{table}
    \centering
    \caption{Distribution of disease classes based on the number of samples.}
    \label{tab:class-distribution}
    \begin{tabular}{>{\raggedright\arraybackslash}p{5.5cm} >{\centering\arraybackslash}p{3.5cm}}
        \toprule
        \textbf{Class Category} & \textbf{Number of Classes} \\
        \midrule
        Common (>10,000 samples) & 11 \\
        Medium (1,000–10,000 samples) & 17 \\
        Rare (<1,000 samples) & 12 \\
        \bottomrule
    \end{tabular}
\end{table}

Furthermore, we split the dataset into training and test sets using an 80:20 ratio, with no patient overlap between the sets. Additionally, we employ 5-fold cross-validation to robustly evaluate model performance. Text is generated from the annotation data provided by MICCAI challenge as in Sec. \ref{sec:text-gen}. 

\subsubsection{Implementation Details}
We use the ViT-B/32 backbone for CLIP \cite{dosovitskiy2021imageworth16x16words} and conduct our experiments on a single node with 1 × NVIDIA RTX 2080 Ti GPU (11 GB memory). All images are scaled to 224x224 in keeping with the original CLIP architecture. The implementation is based on PyTorch version 2.4.0+cu118. We use a learning rate of \( 1e-6 \) and a batch size $bs$ of 32. The AUC scores for base  and rare classes are reported as the average of 5 runs.  We use \( N=40 \) as our dataset has labels for 40 pathologies. During the final loss calculation, both \( \mathcal{L}_c \) and  \( \mathcal{L}_m \) are equally weighted, ensuring a balanced contribution to the optimization process. To optimize training, we employ a ReduceLROnPlateau \cite{Mukherjee2019} learning rate scheduler with a reduction factor of 0.1 and a patience of 2 epochs.


\section{Results}

Table~\ref{tab:performance_1} indicates the 5-fold average AUC scores for all, base and rare classes on the validation set \cite{johnson2019mimiccxrjpglargepubliclyavailable}. Comparisons with other VL SOTA methods are also shown. All the comparison models are trained and evaluated in similar way. MedClip \cite{wang2022medclipcontrastivelearningunpaired}, MedKLIP \cite{wu2023medklipmedicalknowledgeenhanced}, and SLIP \cite{mu2021slipselfsupervisionmeetslanguageimage}, achieve average AUCs of 0.476, 0.576, and 0.502, respectively. MedKLIP performs slightly better on rare classes (0.574 AUC) compared to SLIP (0.495 AUC) and MedClip (0.450 AUC). These models reflect considerable classification uncertainty, particularly for rare findings. 

CheXzero \cite{CheXzero}, our baseline model, achieves a total AUC of 0.644, with 0.647 and 0.631 AUC for base and rare classes, respectively. On the other hand, our method, CXR-CML, achieves a macro AUC of 0.715, with 0.711 and 0.72 AUC for base and rare classes, respectively. These results indicate that CXR-CML successfully achieves SOTA when evaluated over a comprehensive list of categories. This is attributed to our method effectively learning the manifold of data distribution in the CXR dataset. Hence, we observe that model can even discern rarer classes that are less than 2\% of the dataset. 

\begin{table}[t]
    \centering
    \caption{Zero-shot performance comparison of different methods depicted as average AUC for the 28 base classes and 12 rare classes. Results are averaged over 10 runs (std dev. \(< 0.04\)). Statistical significance (\(p < 0.00001\)) is indicated by \(^{*}\) when compared to baseline.}
    \label{tab:performance_1}
    \begin{tabular}{>{\raggedright\arraybackslash}p{3.5cm} >{\centering\arraybackslash}p{2.5cm} >{\centering\arraybackslash}p{2.5cm} >{\centering\arraybackslash}p{2.5cm}}
        \toprule
        \textbf{Method} & \textbf{Total AUC} & \textbf{Base AUC} & \textbf{Rare AUC} \\ 
        \midrule
        MedClip \cite{wang2022medclipcontrastivelearningunpaired} & 0.476 & 0.488 & 0.450 \\
        MedKLIP \cite{wu2023medklipmedicalknowledgeenhanced} & 0.576$^{*}$ & 0.574$^{*}$ & 0.574$^{*}$ \\
        SLIP \cite{mu2021slipselfsupervisionmeetslanguageimage} & 0.502 & 0.505 & 0.495 \\
        CheXzero (Baseline) \cite{CheXzero} & 0.644$^{*}$ & 0.647$^{*}$ & 0.631$^{*}$ \\
        \textbf{CXR-CML (Ours)} & \textbf{0.715}$^{*}$ & \textbf{0.711}$^{*}$ & \textbf{0.720}$^{*}$ \\ 
        \bottomrule
    \end{tabular}
\end{table}

\subsubsection{Ablation Study}

Table ~\ref{tab:auc_results} shows the ablation study starting from the baseline CheXzero , and with different model configurations. The impact of various batch sizes $bs$ and degrees of freedom (\(\nu\)) in Student's t-distribution is shown. The best performance is achieved with a batch size of 32 and \(\nu = 4\), highlighting the importance of careful hyperparameter tuning for optimal results.


\begin{table}[t]
    \centering
    \caption{Ablation study of CXR-CML with different configurations compared to the CheXzero baseline. Results are reported as average AUC for base and rare classes. Statistical significance (\(p < 0.00001\)) is indicated by \(^{*}\). Comparison of AUC scores for different model configurations. $bs$ is batch size and $\nu$ is degree of freedom}
    \label{tab:ablation}
    \begin{tabular}{>{\raggedright\arraybackslash}p{4.0cm} >{\centering\arraybackslash}p{2.5cm} >{\centering\arraybackslash}p{2.5cm} >{\centering\arraybackslash}p{2.5cm}}
        \toprule
        \textbf{Method} & \textbf{Total AUC} & \textbf{Base AUC} & \textbf{Rare AUC} \\ 
        \midrule
        CheXzero (Baseline) \cite{CheXzero} & 0.644$^{*}$ & 0.647$^{*}$ & 0.631$^{*}$ \\
        CheXzero + Meta labels & 0.691$^{*}$ & 0.681$^{*}$ & 0.707$^{*}$ \\
        \midrule
        CXR-CML ($bs = 32$, $\nu = 2$) & 0.710 & 0.710 & 0.710 \\
        CXR-CML ($bs = 32$, $\nu = 4$) & \textbf{0.715}$^{*}$ & \textbf{0.711}$^{*}$ & \textbf{0.720}$^{*}$ \\
        CXR-CML ($bs = 32$, $\nu = 6$) & \textbf{0.715} & 0.710 & 0.718 \\
        \midrule
        CXR-CML ($bs = 16$, $\nu = 4$) & 0.705 & 0.700 & 0.710 \\
        CXR-CML ($bs = 32$, $\nu = 4$) & \textbf{0.715}$^{*}$ & \textbf{0.711}$^{*}$ & \textbf{0.720}$^{*}$ \\
        CXR-CML ($bs = 64$, $\nu = 4$) & 0.711 & 0.707 & 0.716 \\
        \bottomrule 
    \end{tabular}
\end{table}

\paragraph{\textbf{Degrees of Freedom (\(\nu\))}}  
We investigate the impact of the degrees of freedom parameter (\(\nu\)) in Student's t-distribution. We observe that our classification results improve with increasing \(\nu\), while \(\nu\) approaches values defining a Gaussian distribution \cite{ley2014valuemodemultivariatet}. The Gaussian distribution AUC scores are marginally higher, but without statistical significance when compared to \(\nu\)=6. However, our experiments indicate that applying a student t-distribution results in greater model stability.


\paragraph{\textbf{Batch Size}}  
We evaluate the effect of batch size on model performance. A batch size of 32 achieves the best performance, with an overall AUC of 0.715. Increasing the batch size to 64 results in a slight performance drop (0.715 to 0.711 AUC). 
Smaller batch sizes (e.g., 16 and 8) lead to more significant performance degradation (0.715 to 0.705 AUC), likely due to dependence of the contrastive loss $\mathcal{L_c}$ on the batch size \cite{clip}. 

\begin{figure}[h]
    \centering
    \begin{subfigure}[b]{0.45\textwidth}
        \centering
        \includegraphics[width=\linewidth]{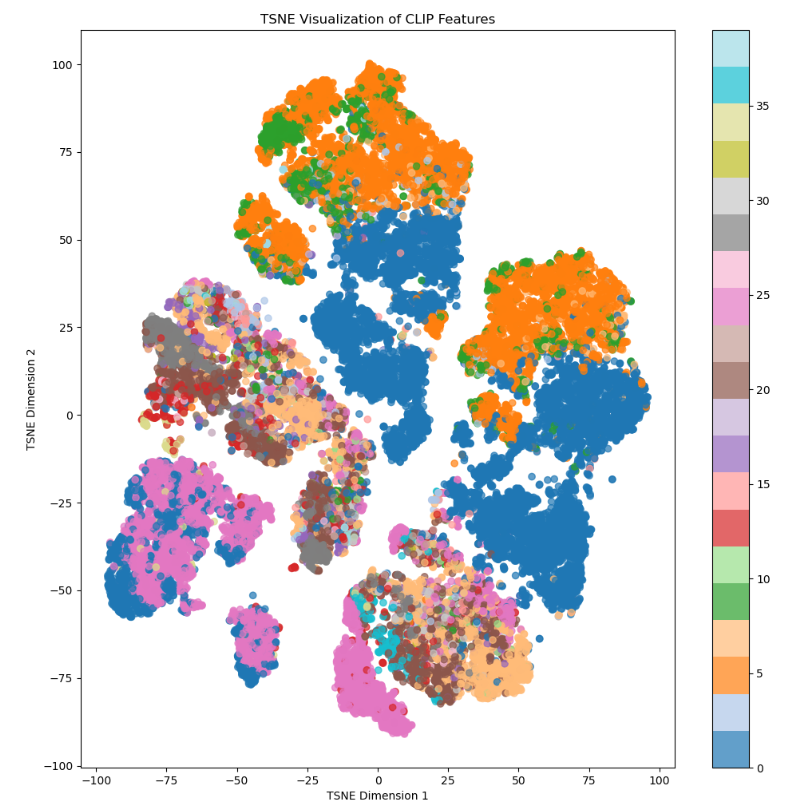}
        \caption{Latent space of CheXzero}
        \label{fig:main-latent} 
    \end{subfigure}
    \hfill
    \begin{subfigure}[b]{0.45\textwidth}
        \centering
        \includegraphics[width=\linewidth]{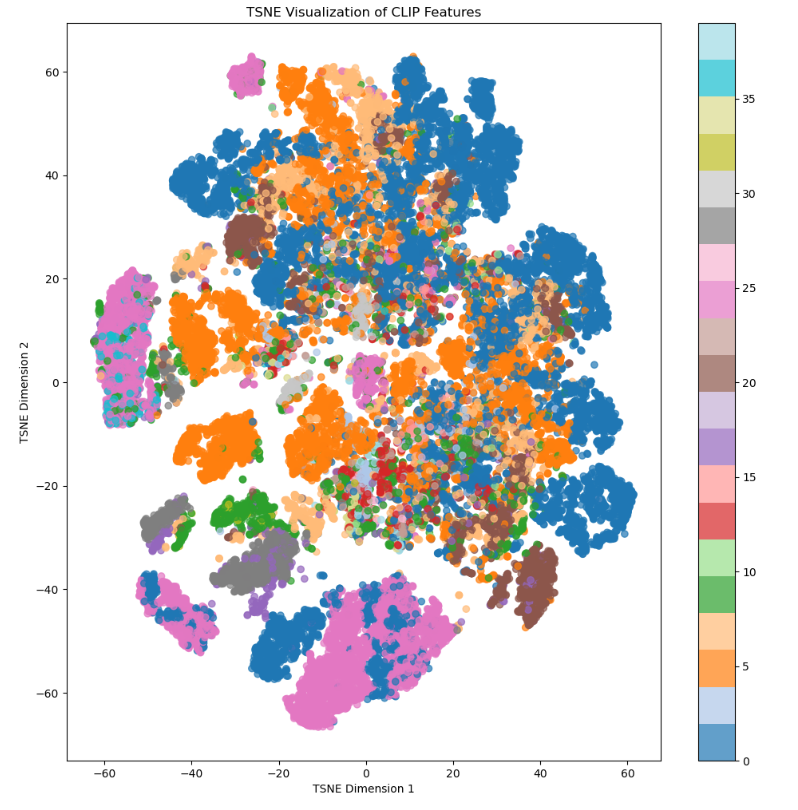}
        \caption{Latent space of CXR-CML}
        \label{fig:latent-2}
    \end{subfigure}
    \vspace{-3mm}  
    \caption{ t-SNE visualization of CLIP features for different models. The plot represents the latent space clustering of 40 disease classes using CheXzero and CXR-CML over full validation set.}
    \label{fig:side-by-side}
\end{figure}

\paragraph{\textbf{Feature visualization}} 
Fig. \ref{fig:side-by-side} illustrates the t-SNE plots of both the CheXzero baseline and our method, with a batch size of 32. Though the baseline results in well defined clusters, the quantitative results show our method demonstrates better performance across the complete list of categories. Overall the results suggest that CXR-CML can accurately capture the differentiating characteristics of the long-tailed classes.

\section{Conclusion}

CXR-CML showcases a notable enhancement in zero-shot classification performance, by modeling the latent space with an emphasis on the long-tailed data distribution. In this study, we employ the Student t-distribution to provide a robust mathematical framework for clustering, enhancing the representation of underrepresented categories. This improved clustering further strengthens the subsequent metric learning stage, leading to enhanced classification performance. Future work will explore additional VL-SSL methods as comparitive baselines and extend evaluation to other medical domains.

\clearpage
%
%
\bibliography{ref}
\bibliographystyle{splncs04}
\end{document}